# Hierarchically Learned View-Invariant Representations for Cross-View Action Recognition

Yang Liu, Zhaoyang Lu, *Senior Member, IEEE*, Jing Li, *Member, IEEE*, and Tao Yang, *Member, IEEE*

*Abstract*—Recognizing human actions from varied views is challenging due to huge appearance variations in different views. The key to this problem is to learn discriminant view-invariant representations generalizing well across views. In this paper, we address this problem by learning view-invariant representations hierarchically using a novel method, referred to as joint sparse representation and distribution adaptation. To obtain robust and informative feature representations, we first incorporate a sample-affinity matrix into the marginalized Stacked Denoising Autoencoder to obtain shared features that are then combined with the private features. In order to make the feature representations of videos across views transferable, we then learn a transferable dictionary pair simultaneously from pairs of videos taken at different views to encourage each action video across views to have the same sparse representation. However, the distribution difference across views still exists because a unified subspace, where the sparse representations of one action across views are the same, may not exist when the view difference is large. Therefore, we propose a novel unsupervised distribution adaptation method that learns a set of projections that project the source and target views data into respective low-dimensional subspaces, where the marginal and conditional distribution differences are reduced simultaneously. Therefore, the finally learned feature representation is view-invariant and robust for substantial distribution difference across views even though the view difference is large. Experimental results on four multi-view datasets show that our approach outperforms the state-of-the-art approaches.

## I. INTRODUCTION

HUMAN action recognition aims to automatically recognize an ongoing action from a video clip, which has received great attention in recent years due to its wide applications, including video surveillance [1], video labeling [2], video content retrieval [3], human-computer interaction [4], and sports video analysis [5]. However, recent works

Manuscript received November 23, 2017; revised March 7, 2018, April 10, 2018, and June 4, 2018; accepted August 22, 2018. Date of publication August 31, 2018; date of current version August 2, 2019. This work was supported by the National Natural Science Foundation of China under Grant 61502364 and Grant 61672429. This paper was recommended by Associate Editor Y. Wu. *(Corresponding author: Jing Li.)*

Y. Liu, Z. Lu, and J. Li are with the School of Telecommunications Engineering, Xidian University, Xi'an 710071, China (e-mail: yliu_0@stu.xidian.edu.cn; zhylu@xidian.edu.cn; jinglixd@mail.xidian.edu.cn).

T. Yang is with the School of Computer Science, Northwestern Polytechnical University, Xi'an 710072, China (e-mail: tyang@nwpu.edu.cn).

Color versions of one or more of the figures in this paper are available online at http://ieeexplore.ieee.org.

Digital Object Identifier 10.1109/TCSVT.2018.2868123

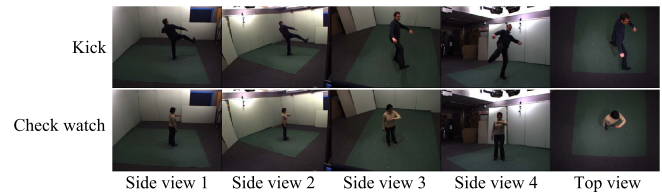

Fig. 1. Examples of multi-camera-view on IXMAS dataset.

in [6]–[11] have demonstrated that recognizing action data in cross-view scenario is challenging due to large appearance variations in action videos captured by various cameras at different locations. For example, the same action performed by the same actor may be visually different from one view to another view (Figure 1). In addition, different viewpoints of cameras may result in different background, camera motions, lighting conditions and occlusions. Therefore, developing methods for cross-view action recognition that can recognize an unknown action in the target view by using the features extracted from some other source views remains a challenge.

In order to accurately recognize human actions from varied views, a family of view-shared sparse representation based approaches are proposed recently and demonstrated to achieve good results [12]–[14]. They assume that samples from different views contribute equally in shared features and ignore view-private features. However, this assumption is not always valid, i.e, the top view should have lower contribution to the shared features compared to other side views (e.g. the top view and side views in Figure 1). Actually, shared features of one action across views mainly encode the body and body outline while private features mainly encode different limb poses that represent the class information across views [14]. Therefore, the view-private features that capture motion information particularly owned by one view should be incorporated into view-shared features to learn more discriminative and informative features. In addition, the distribution difference across views still exists because a unified subspace where the feature representations of one action across views are the same may not exist when the view difference is large (e.g. the top view and the side views in Figure 1). This will degrade the overall performance of the cross-view action recognition algorithm. Thus, we should learn a set of projections that project different views into respective subspaces to obtain new representations of respective views, and concurrently encourage the subspace divergence to be small. In this way, the learned representations can generalize well across views even when the view difference is large.

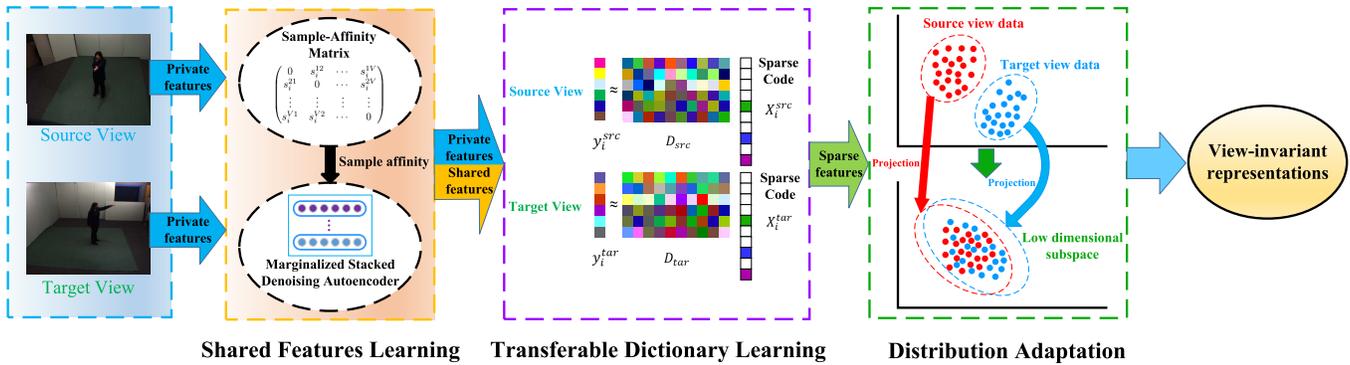

Fig. 2. Framework of our proposed JSRDA. The framework is hierarchical as the view-invariant representation is learned in a coarse-to-fine fashion.

In this paper, we focus on unsupervised cross-view action recognition problem including the case when the view difference is large and learn view-invariant representations hierarchically by incorporating shared features learning, transferable dictionary learning and distribution adaptation into a unified framework named Joint Sparse Representation and Distribution Adaptation (JSRDA). An overview of the JSRDA is present in Figure 2. The proposed method JSRDA mainly consist of three stages.

In the first stage, a Sample-Affinity Matrix (SAM) introduced in [15] is employed to measure the similarities between video samples in different views, which facilitates accurately balancing information transfer across views. Then the SAM is incorporated into the marginalized stacked denoising Autoencoder [16] (mSDA) to learn more robust shared features. In addition to shared features, private features that originated from raw input features are combined with the obtained shared features to yield more informative feature representations.

In the second stage, we learn a set of dictionaries that correspond to training and testing views respectively. These dictionaries are learned simultaneously from the sets of videos taken at different views by encouraging each video in the set to have the same sparse representations. After the dictionaries are learned, we obtain the sparse representations of training and testing videos respectively using the corresponding dictionary. This procedure enables the transfer of sparse feature representations of videos in the source view(s) to the corresponding videos in the target view.

However, the distribution difference across views still exists after the second stage because a unified subspace where the sparse representations of one action across views are the same may not exist when the view difference is large (e.g. the top view and the side views). Therefore, in the third stage, we propose a novel unsupervised distribution adaptation method that learns a set of projections that project the source and target views into respective subspaces where both the marginal and conditional distribution differences between source and target views are reduced simultaneously. After the projections, 1) the variance of target view data is maximized to preserve the embedded data properties, 2) the discriminative information of source view data is preserved to effectively transfer the class information, 3) both the marginal and conditional distribution differences between source and target views are minimized, 4) the divergence of these projections is encouraged to be small to reduce the domain divergence between source and target views.

Finally, the view-invariant representations of action videos from different views are obtained in their respective subspaces. Then, we train a classifier in the source view(s) and test it in the target view. Extensive experiments on four multi-view datasets show that our approach significantly outperforms state-of-the-art approaches.

The main contributions of this paper are as follows:

- To obtain more robust and informative feature representations for cross-view action recognition, a sample-affinity matrix is incorporated into the marginalized stacked denoising Autoencoder (mSDA) to learn shared features, which are then combined with private features.
- To address the performance degradation problem when the view difference is large, we propose a novel unsupervised distribution adaptation method that learns a set of projections that project the source and target views into respective subspaces where both the marginal and conditional distribution differences between source and target views are reduced simultaneously.
- To obtain view-invariant feature representations that generalize well across views, we learn view-invariant representations hierarchically by incorporating shared features learning, transferable dictionary learning and distribution adaptation into a unified framework, which is effective and can learn robust and discriminative view-invariant representations even on different datasets.

This paper is organized as follows: Section II briefly reviews related state-of-the-art works. Section III introduces the proposed approach JSRDA for cross-view action recognition. Experimental results and related discussions are presented in Section IV. Finally, Section V concludes the paper.

## II. RELATED WORK

### A. Cross-View Action Recognition

Recently, many approaches have been proposed to address the problem of cross-view action recognition.

Farhadi and Tabrizi [17] employed maximum margin clustering (MMC) to generate split-based features of source view and then transferred the split values to the target view. Zhang et al. [18] added a temporal regularization on the traditional MMC. These works require feature-to-feature correspondence at the frame-level. Liu et al. [19] presented a bipartite-graph-based method to bridge the domain shift across view-dependent vocabularies. Zheng et al. [14] exploited the video-to-video correspondence and proposed a dictionary learning based method to jointly learn a set of view-specific dictionaries for specific views and a common dictionary shared across different views. Li and Zickler [7] proposed "virtual views" that connect the source and target views by a virtual path, which is associated with a linear transformation of the action descriptors. Similarly, Zhang et al. [8] intended to bridge the source view and the target view by a continuous virtual path keeping all the visual information. Wang et al. [20] proposed a statistical translation framework by estimating the visual word transfer probabilities across views for cross-view action recognition. Kong et al. [15] addressed the cross-view action recognition problem by learning view-specific and view-shared features using a marginalized autoencoder based deep models. Yan et al. [21] proposed a Multi-Task Information Bottleneck (MTIB) clustering method to explore the shared information between multiple action clustering tasks to improve the performance of individual task. Ulhaq et al. [22] proposed an advanced space-time filtering framework for recognizing human actions despite large viewpoint variations. Rahmani et al. [23] proposed a 3D human model based cross-view action recognition method that learns a single model to transform any action from any viewpoints to its respective high level representation without requiring action labels or knowledge of the viewing angles.

Different from above-mentioned cross-view action recognition approaches [7], [8], [14], [15], [17]–[23], our proposed approach exploits both the view-private and view-shared features to learn view-invariant representations hierarchically by incorporating shared feature learning, transferable dictionary learning and distribution adaptation into a unified cross-view action recognition framework. To address the problem of performance degradation when view difference is large, we learn a set of projections that project the source and target views into respective subspaces to make the learned representation generalize well across views. In this way, our approach can learn robust and discriminative view-invariant representations for cross-view action recognition even with large view difference or different datasets.

### B. Transfer Learning on Heterogeneous Features

From the perspective of transfer learning, our work is related to the subspace based methods [24]–[27]. Wang and Mahadevan [24] proposed a manifold alignment based method to learn a common feature subspace for all heterogeneous domains by preserving the topology of each domain, matching instances with the same labels and separating instances with different labels. However, this method needs class labels of both source and target domains and requires that the data should have a manifold structure, while our method is unsupervised and does not require the manifold assumption of data. Long et al. [25] proposed a distribution adaptation method to find a common subspace where the marginal and conditional distribution shifts between domains are reduced. Zhang et al. [26] relaxed the assumption that there exists a unified projection to map source and target domains into a unified subspace, they learned two projections to map the source and target domains into their respective subspaces where both the geometrical and statistical distribution difference are minimized. Long et al. [27] introduced a unsupervised domain adaptation method that reduces the domain shift by jointly finding a common subspace and reweighting the instances across domains.

Different from previous distribution adaptation methods [24]–[27], we do not assume that there exists a unified projection since this assumption is invalid when the distribution difference across views is large. Instead, we learn a set of projections that project the source and target views into respective subspaces to obtain new representations of respective views, and concurrently encourage the subspace divergence across views to be small.

## III. JOINT SPARSE REPRESENTATION AND DISTRIBUTION ADAPTATION

The purpose of this work is to learn view-invariant representations that allow us to train a classifier on one (or multiple) view(s), and test on the other view.

### A. Shared Features Learning

This subsection aims to obtain new informative feature representations of action videos for further transferable dictionary learning and distribution adaptation.

*1) Sample-Affinity Matrix (SAM):* To fix notation, we consider training videos of V views: $\{X^v, \mathbf{y}^v\}_{v=1}^V$. The data instances of the $v$-th view $X^v$ consist of $N$ action videos: $X^v = [\mathbf{x}_1^v, \cdots, \mathbf{x}_N^v] \in \mathbb{R}^{d \times N}$ with corresponding labels $\mathbf{y}^v = [y_1^v, \cdots, y_N^v]$, where $\mathbf{x}_i^v$ ($i = 1, \cdots, N$) denotes the feature of the video $i$ of the $v$-th view and $d$ denotes the dimensionality of the video feature. We employ the Sample-Affinity-Matrix (SAM) introduced in [15] to measure the similarity between pairs of video samples in multiple views. The SAM $S \in \mathbb{R}^{VN \times VN}$ is defined as a block diagonal matrix:

$$S = \text{diag}(S_1, \cdots, S_N), \quad S_i = \begin{pmatrix} 0 & s_i^{12} & \cdots & s_i^{1V} \\ s_i^{21} & 0 & \cdots & s_i^{2V} \\ \vdots & \vdots & \vdots & \vdots \\ s_i^{V1} & s_i^{V2} & \cdots & 0 \end{pmatrix},$$

where $\text{diag}(\cdot)$ creates a diagonal matrix, and $s_i^{uv} = \exp(\|\mathbf{x}_i^v - \mathbf{x}_i^u\|^2 / 2c)$ parameterized by $c$ calculates the distance of the $i$-th video sample between two views. In this paper, we use $c = 2$ in all the experiments according to the default setting in [28].

Actually, SAM S captures both intra-class between-view information and between-class intra-view information. A block $S_i$ in S tells us how an action varies if view changes because it characterizes appearance variations in different views within one class, which allows us to transfer information between views and learn robust cross-view features. In addition, the off-diagonal blocks in SAM S are set to zeros to limit information sharing between classes in the same view. As a result, the features from different classes but in the same view are encouraged to be distinct, which enables us to differentiate various action classes if they appear similarly in some views.

*2) Autoencoders:* Our shared features learning approach builds upon a popular deep learning approach Autoencoder (AE) [29] and a AE based domain adaptation method marginalized stacked denoising Autoencoder [16] (mSDA). The objective of AE is to encourage similar or identical input-output pairs where the reconstruction loss is minimized. In this way, the hidden unit is a good representation of the inputs as the reconstruction process captures the intrinsic structure of the input data. Different from the two-level encoding and decoding in AE, marginalized stacked denoising Autoencoder (mSDA) learns robust data representation using a single mapping $W = \arg\min_W \sum_{i=1}^N \|\mathbf{x}_i - W\tilde{\mathbf{x}}_i\|^2$ by recovering original features from data that are artificially corrupted with noise, where $\tilde{\mathbf{x}}_i$ is the corrupted version of $\mathbf{x}_i$ obtained by randomly setting each feature to 0 with a probability $p$, and $N$ is the number of training samples. mSDA performs $m$ times over the training set with different corruptions each time, which essentially performs a dropout regularization on the mSDA [30]. By setting $m \to \infty$, mSDA effectively uses infinitely many copies of noisy data to compute the mapping matrix $W$ that is robust to noise. mSDA is stackable and can be computed in closed-form.

*3) Single-Layer Shared Features Learning:* Actually, an action from one view has some similar appearance information with that from other views. This motivates us to reconstruct an action data from one view (target view) using the action data from other view(s) (source view(s)). In this way, shared information between views can be refined and transferred to the target view. Inspired by the mSDA, we incorporate SAM S into the mSDA to balance information transfer between views and learn discriminative shared features across multiple views. We learn shared features using the following objective function that define the discrepancy between data of the $v$-th target view and the data of all the $V$ source views:

$$\arg\min_W \psi = \sum_{i=1}^N \sum_{v=1}^V \|W\tilde{\mathbf{x}}_i^v - \sum_u \mathbf{x}_i^u s_i^{uv}\|^2$$
$$= \|W\tilde{X} - XS\|_F^2, \quad (1)$$

where $s_i^{uv}$ is a weight measuring the contributions of the $u$-th view action in the reconstruction of the action sample $\mathbf{x}_i^v$ of the $v$-th view. $W \in \mathbb{R}^{d \times d}$ is the mapping matrix for the corrupted input $\tilde{\mathbf{x}}_i^v$ of all the views. $S \in \mathbb{R}^{VN \times VN}$ is a sample-affinity matrix encoding all the weights $\{s_i^{uv}\}$. Matrices $X$, $\tilde{X} \in \mathbb{R}^{d \times VN}$ denote the input training matrix and the corresponding corrupted version of $X$, respectively [16].

The solution to optimization problem in Eq. (1) can be expressed as the well-known closed-form solution for ordinary least squares [16], [31]:

$$W = (XS\tilde{X}^T)(\tilde{X}\tilde{X}^T)^{-1} \quad (2)$$

It should be noted that $XS\tilde{X}^T$ and $\tilde{X}\tilde{X}^T$ are computed by repeating the corruption $m \to \infty$ times. By the weak law of large numbers [16], $XS\tilde{X}^T$ and $\tilde{X}\tilde{X}^T$ can be computed by their expectations $E_p(XS\tilde{X}^T)$ and $E_p(\tilde{X}\tilde{X}^T)$ with the corruption probability $p$, respectively.

Although the mSDA can be designed to have deep architecture by layer-wise stacking, we use only one layer in this paper considering the extra training time using multiple layers. To obtain the shared features, a nonlinear squashing function $\sigma(\cdot)$ is applied on the output of one layer: $H_s = \sigma(WX)$, where $X$ denotes the raw features of training data and $H_s$ denotes the shared features. Throughout this paper, we use $\tanh(\cdot)$ as the squashing function. Besides the information shared across views, private features that capture discriminative information exclusively exists in each view should also be taken into consideration. Therefore, original features $X$ is considered as private features $H_p$ and concatenated with the obtained shared feature $H_s$ to form the new informative representation $H_{sp} = [H_s, H_p] \in \mathbb{R}^{2d \times N}$.

### B. Transferable Dictionary Learning

Although the obtained new representation $H_{sp}$ contains both shared and private features, it cannot capture view-invariant information due to the variations in feature representations of the same action from different views. Therefore, we employ transferable dictionary learning method introduced in [14] to learn sparse representation for each action video based on the new representation $H_{sp}$. Specifically, we learn a set of view-specific dictionaries where each dictionary corresponds to one camera view. These dictionaries are learned simultaneously from the sets of corresponding videos taken at different views with the aim to encourage each video in the set to have the same sparse representation. In this way, videos of the same action class from source and target views will tend to have the same sparse codes when reconstructed from the corresponding view-specific dictionary.

In this paper, we consider unsupervised transferable dictionary learning where labels of correspondence videos are not available. In addition, we require that the number of training actions videos in each view should be the same. Suppose there are $p$ source views and one target view. To be noticed, the cross-view problem is a special case of multi-view problem when $p = 1$. Let $D_{s,i} \in \mathbb{R}^{2d \times K}$ and $D_t \in \mathbb{R}^{2d \times K}$ denote the view-specific dictionary of the $i$-th source view and the target view, respectively. $K$ is the number of dictionary atoms and each view-specific dictionary is of the same size. $Y_{s,i} \in \mathbb{R}^{2d \times N}$ and $Y_t \in \mathbb{R}^{2d \times N}$ denote the feature representation of the $i$-th source view and the target view, respectively. The sparse representations $X \in \mathbb{R}^{K \times N}$ are obtained by solving the following objective function:

$$\arg\min_{\{D_{s,i}\}_{i=1}^p, D_t, X} \sum_{i=1}^p \|Y_{s,i} - D_{s,i}X\|_2^2 + \|Y_t - D_t X\|_2^2$$
$$s.t. \ \forall i, \ \|x_i\|_0 \leq \gamma \quad (3)$$

Since we have the same number of action videos in each view, Eq. (3) can be rewritten as:

$$\arg\min_{D,X} \|Y - DX\|_2^2$$
$$s.t. \ \forall i, \quad \|x_i\|_0 \le \gamma \quad (4)$$

where $Y = \begin{bmatrix} Y_{s,1} \\ \cdots \\ Y_{s,p} \\ Y_t \end{bmatrix}$, $D = \begin{bmatrix} D_{s,1} \\ \cdots \\ D_{s,p} \\ D_t \end{bmatrix}$ and $\|x_i\|_0 \le \gamma$ is the sparsity constraint. As for the optimization of the view-specific dictionaries $D$, they can be learned by the K-SVD [32] algorithm. After obtaining these dictionaries, the OMP [33] algorithm can be employed to compute the sparse feature representations. Consequently, all videos in all views are projected into a unified view-invariant sparse feature space. This procedure enables the transfer of sparse feature representations of videos in the source view(s) to the corresponding videos in the target view.

### C. Distribution Adaptation

Although the obtained sparse representations of one action in all views are the same, the distribution difference across views still exists because a unified subspace where the sparse representations of one action across views are the same may not exist when the view difference is large (e.g. the top view and the side views). This will degrade the overall performance of the cross-view action recognition algorithm. Thus, we relax this strong assumption that there exists a unified subspace where the feature representations of one action in all views should be strictly equal. Instead, we learn a set of projections that project different views into respective subspaces to obtain new representations of respective views, and concurrently encourage the subspace divergence to be small.

*1) Problem Definition:* Suppose there are $p$ source views and one target view with a total of $C$ classes. To fix the definitions of terminologies, the data from the $i$-th source view denoted as $X_{s,i} \in \mathbb{R}^{K \times N_{s,i}}$ are draw from marginal distribution $P_{s,i}(X_{s,i})$ and the target view data $X_t \in \mathbb{R}^{K \times N_t}$ are draw from marginal distribution $P_t(X_t)$, where $K$ is the dimension of the data instance, $N_{s,i}$ and $N_t$ are the number of samples in the $i$-th source view and the target view, respectively. In unsupervised distribution adaptation, there are sufficient labeled source view data and unlabeled target domain data in the training stage. We assume that the features and label spaces between source and target views are the same. Due to the domain divergence between views, for any $i \in \{1, \cdots, p\}$, marginal distribution $P_{s,i}(X_{s,i}) \ne P_t(X_t)$ and conditional distribution $P_{s,i}(Y_{s,i} \mid X_{s,i}) \ne P_t(Y_t \mid X_t)$, where $Y_{s,i} \in \mathbb{R}^{1 \times N_{s,i}}$ and $Y_t \in \mathbb{R}^{1 \times N_t}$ are the class labels of the $i$-th source view data and the target view data, respectively. Different from previous distribution adaptation methods, we do not assume that there exists a unified transformation T that $P_{s,i}(\text{T}(X_{s,i})) = P_t(\text{T}(X_t))$ and $P_{s,i}(Y_{s,i} \mid \text{T}(X_{s,i})) = P_t(Y_t \mid \text{T}(X_t))$, since this assumption becomes invalid when the distribution shift across views is large. Instead, we propose a novel distribution adaptation method that learns a set of projections that project the source and target views into respective subspaces to obtain new representations of respective views, and encourage the subspace divergence to be small at the same time.

*2) Formulation:* Our proposed distribution adaptation approach is formulated by finding a set of projections ($F_{s,i}$ for the $i$-th source view and $F_t$ for the target view) to obtain new representations of respective views, such that 1) the difference in marginal distribution and conditional distribution across views is small, 2) the divergence between source and target subspaces is small, 3) the variance of target view domain is maximized, 4) the discriminative information of source view domain is preserved.

To reduce the difference between the marginal distributions $P_{s,i}(X_{s,i})$ and $P_t(X_t)$, we follow [25] and [34]–[36] and employ the empirical Maximum Mean Discrepancy (MMD) to compute the distance between the sample means of the source and target data in the k-dimensional embeddings,

$$\min_{\{F_{s,i}\}_{i=1}^{p}, F_t} \sum_{i=1}^{p} \|\frac{1}{N_{s,i}} \sum_{x_k \in X_{s,i}} F_{s,i}^{\text{T}} x_k - \frac{1}{N_t} \sum_{x_j \in X_t} F_t^{\text{T}} x_j\|_F^2 \quad (5)$$

In order to reduce difference between the conditional distributions $P_{s,i}(Y_{s,i} \mid X_{s,i})$ and $P_t(Y_t \mid X_t)$, sufficient labeled data in target view is need. However, there are no labeled data in the target view in unsupervised scenario. To address these issues, Long *et al.* [25] utilized target view pseudo labels predicted by source view classifier to represent the conditional distribution in the target view domain. The pseudo labels in target view domain are iteratively refined to reduce the difference in conditional distributions with the source view domains. We follow this idea to minimize the conditional distribution difference between domains,

$$\min_{\{F_{s,i}\}_{i=1}^{p}, F_t} \sum_{i=1}^{p} \sum_{c=1}^{C} \|\frac{1}{N_{s,i}^{(c)}} \sum_{\mathbf{x}_k \in X_{s,i}^{(c)}} F_{s,i}^{\text{T}} \mathbf{x}_k - \frac{1}{N_t^{(c)}} \sum_{\mathbf{x}_j \in X_t^{(c)}} F_t^{\text{T}} \mathbf{x}_j\|_F^2$$
$$(6)$$

where $X_{s,i}^{(c)}$ is the set of data instances from class $c$ in the $i$-th source view and $N_{s,i}^{(c)}$ is the number of data instances in $X_{s,i}^{(c)}$. Correspondingly, $X_t^{(c)}$ is the set of data instances from class $c$ in the target view and $N_t^{(c)}$ is the number of data instances in $X_t^{(c)}$. Since we have the same number of action videos in each view, the marginal distribution difference minimization term Eq. (5) and conditional distribution difference minimization term Eq. (6) can be combined to obtain the final distribution divergence minimization term,

$$\min_{F_s, F} Tr([F_s^{\text{T}} \quad F^{\text{T}}] \begin{bmatrix} M_s & M_{st} \\ M_{ts} & M \end{bmatrix} \begin{bmatrix} F_s \\ F \end{bmatrix}) \quad (7)$$

where the formulation of $F_s$, $F$, $M_s$, $M_{st}$, $M_{ts}$ and $M$ can be found in Appendix A.

To reduce the divergence between source and target subspaces, we use the following term to encourage the source and target subspaces to be close,

$$\min_{\{F_{s,i}\}_{i=1}^{p}, F_t} \sum_{i=1}^{p} \|F_{s,i} - F_t\|_F^2 \quad (8)$$

We rewrite Eq. (8) as follows,

$$\min_{F_s, F} \|F_s - F\|_F^2 \quad (9)$$

where $F_s = \begin{bmatrix} F_{s,1} \\ \cdots \\ F_{s,p} \end{bmatrix}$ and $F = \begin{bmatrix} F_t \\ \cdots \\ F_t \end{bmatrix}$ is obtained by replicating $F_t$ $p$ times.

To maximize the variance of target view data and preserve its embedded data properties, we use the following term to achieve this purpose,

$$\max Tr(F^T S F) \quad (10)$$

where $S = [S_t, \cdots, S_t]$ is obtained by replicating $S_t$ $p$ times, $S_t = X_t H_t X_t^T$ is essentially a covariance matrix, and $H_t = I_t - \frac{1}{N_t} 1_t 1_t^T$ is the centering matrix while $1_t \in \mathbb{R}^{N_t \times 1}$ is the column vector with all ones and $I_t \in \mathbb{R}^{N_t \times N_t}$ is the identity matrix.

Since the label information in the source views is available, we can utilize this to preserve the discriminative information in source views. Therefore, we use following terms to achieve this purpose,

$$\max_{\{F_{s,i}\}_{i=1}^p} \sum_{i=1}^p Tr(F_{s,i}^T S_{b,i} F_{s,i}) \quad (11)$$

$$\min_{\{F_{s,i}\}_{i=1}^p} \sum_{i=1}^p Tr(F_{s,i}^T S_{\omega,i} F_{s,i}) \quad (12)$$

where $S_{b,i}$ is the inter-class variance matrix of the data from the $i$-th source view domain and $S_{\omega,i}$ is the intra-class variance matrix, which are defined as follows,

$$S_{b,i} = \sum_{c=1}^C N_{s,i}^{(c)} (m_{s,i}^{(c)} - \bar{m}_{s,i})(m_{s,i}^{(c)} - \bar{m}_{s,i})^T \quad (13)$$

$$S_{\omega,i} = \sum_{c=1}^C X_{s,i}^{(c)} H_{s,i}^{(c)} (X_{s,i}^{(c)})^T \quad (14)$$

where $X_{s,i}^{(c)} \in \mathbb{R}^{K \times N_{s,i}^{(c)}}$ is the set of data instance from class $c$ in the $i$-th source view, $m_{s,i}^{(c)} = \frac{1}{N_{s,i}^{(c)}} \sum_{k=1}^{N_{s,i}^{(c)}} x_k^{(c)}$, $\bar{m}_{s,i} = \frac{1}{N_{s,i}} \sum_{k=1}^{N_{s,i}} x_k$, $H_{s,i}^{(c)} = I_{s,i}^{(c)} - \frac{1}{N_{s,i}^{(c)}} 1_{s,i}^{(c)} (1_{s,i}^{(c)})^T$ is the centering matrix of data from class $c$, $I_{s,i}^{(c)} \in \mathbb{R}^{N_{s,i}^{(c)} \times N_{s,i}^{(c)}}$ is the identity matrix. $1_{s,i}^{(c)} \in \mathbb{R}^{N_{s,i}^{(c)} \times 1}$ is the column vector with all ones, $N_{s,i}^{(c)}$ is the number of data from class $c$ in the $i$-th source view. Similarly, Eq. (11) and Eq. (12) can be rewritten as follows,

$$\max_{F_s} Tr(F_s^T S_b F_s) \quad (15)$$

$$\min_{F_s} Tr(F_s^T S_\omega F_s) \quad (16)$$

where $F_s = \begin{bmatrix} F_{s,1} \\ \cdots \\ F_{s,p} \end{bmatrix}$, $S_b = [S_{b,1}, \cdots, S_{b,p}]$ and $S_\omega = [S_{\omega,1}, \cdots, S_{\omega,p}]$.

We formulate our distribution adaptation method by incorporating the above five terms Eq. (7), (9), (10), (15) and (16) into a unified objective function as follows:

$$\max \frac{\mu\{\text{T-Var}\} + \beta\{\text{S-Inter-Var}\}}{\{\text{Dis-Dif}\} + \lambda\{\text{Sub-Div}\} + \beta\{\text{S-Intra-Var}\} + \mu\{F\text{-C}\}}$$

where T-Var, S-Inter-Var, Dis-Dif, Sub-Div, S-Intra-Var and $F$-C terms denote the target view data variance, source view inter-class variance, distribution difference, subspace divergence, source view intra-class variance and the scale constraint of $F$, respectively. And $\lambda, \mu, \beta$ are the parameters to balance the importance of each terms. We follow [37] to impose a constraint that $Tr(F^T F)$ is small to control the scale of $F$. Specifically, we aim at finding a set of projections $F_s$ and $F$ by solving the following optimization problem,

$$\max_{F_s, F} \frac{Tr([F_s^T \ F^T] \begin{bmatrix} \beta S_b & 0 \\ 0 & \mu S \end{bmatrix} \begin{bmatrix} F_s \\ F \end{bmatrix})}{Tr([F_s^T \ F^T] \begin{bmatrix} M_s + \lambda I + \beta S_\omega & M_{st} - \lambda I \\ M_{ts} - \lambda I & M + (\lambda + \mu)I \end{bmatrix} \begin{bmatrix} F_s \\ F \end{bmatrix})} \quad (17)$$

where $I$ is the identity matrix.

Actually, minimizing the denominator of Eq. (17) encourages small difference of marginal and conditional distributions, small subspace divergence between the source and target views, and small intra-class variance of the source view. Maximizing the numerator of Eq. (17) encourages the large variance of the target view and large inter-class variance of source views. In addition, we also iteratively update the pseudo labels of target view domain data using the learned transformations to improve the labelling quality until convergence.

*3) Optimization:* To optimize Eq. (17), we rewrite $[F_s^T \ F^T]$ as $W^T$. Then the objective function can be rewritten as follows:

$$\max_W \frac{Tr(W^T \begin{bmatrix} \beta S_b & 0 \\ 0 & \mu S \end{bmatrix} W)}{Tr(W^T \begin{bmatrix} M_s + \lambda I + \beta S_\omega & M_{st} - \lambda I \\ M_{ts} - \lambda I & M + (\lambda + \mu)I \end{bmatrix} W)} \quad (18)$$

Note that the objective function Eq. (18) is invariant to rescaling of $W$. Therefore, the objective function Eq. (18) can be rewritten as follows:

$$\max_W Tr(W^T \begin{bmatrix} \beta S_b & 0 \\ 0 & \mu S \end{bmatrix} W)$$

$$s.t. \ Tr(W^T \begin{bmatrix} M_s + \lambda I + \beta S_\omega & M_{st} - \lambda I \\ M_{ts} - \lambda I & M + (\lambda + \mu)I \end{bmatrix} W) = 1 \quad (19)$$

According to the constrained optimization theory, we denote $\Phi = \text{diag}(\phi_1, \cdots, \phi_k) \in \mathbb{R}^{k \times k}$ as the Lagrange multiplier and derive the Lagrange function for Eq. (19) as:

$$L = Tr(W^T \begin{bmatrix} \beta S_b & 0 \\ 0 & \mu S \end{bmatrix} W)$$

$$+ Tr((W^T \begin{bmatrix} M_s + \lambda I + \beta S_\omega & M_{st} - \lambda I \\ M_{ts} - \lambda I & M + (\lambda + \mu)I \end{bmatrix} W - I)\Phi \quad (20)$$

Setting $\frac{\partial L}{\partial W} = 0$, we obtain the problem of generalized eigendecomposition,

$$\begin{bmatrix} \beta S_b & 0 \\ 0 & \mu S \end{bmatrix} W = \begin{bmatrix} M_s + \lambda I + \beta S_\omega & M_{st} - \lambda I \\ M_{ts} - \lambda I & M + (\lambda + \mu) I \end{bmatrix} W \Phi \quad (21)$$

where $\Phi = \text{diag}(\phi_1, \cdots, \phi_k)$ are the $k$ smallest eigenvectors. Finally, finding the optimal transformation matrix $W$ is reduced to solve Eq. (21) for the $k$ smallest eigenvectors $W = [W_1, \cdots, W_k]$. Once the transformation matrix $W$ is obtained, the corresponding subspace projections $F_s = \begin{bmatrix} F_{s,1} \\ \cdots \\ F_{s,p} \end{bmatrix}$ and $F_t$ can be easily obtained.

The distribution adaptation method can be extended to nonlinear problems in a Reproducing Kernel Hilbert Space (RKHS) using kernel mapping $\psi : \mathbf{x} \mapsto \psi(\mathbf{x})$, or $\psi(X) = [\psi(\mathbf{x}_1), \cdots, \psi(\mathbf{x}_N)]$, and kernel matrix $K = \psi(X)^T \psi(X) \in \mathbb{R}^{N \times N}$, where $N$ is the number of all samples in source and target views. We use the Representer theorem to kernelized the objective function Eq. (17) as follows:

$$\max_{F_s, F} \frac{Tr([F_s^T \ F^T] \begin{bmatrix} \beta S_b & 0 \\ 0 & \mu S \end{bmatrix} \begin{bmatrix} F_s \\ F \end{bmatrix})}{Tr([F_s^T \ F^T] \begin{bmatrix} M_s + \lambda K + \beta S_\omega & M_{st} - \lambda K \\ M_{ts} - \lambda K & M + (\lambda + \mu) K \end{bmatrix} \begin{bmatrix} F_s \\ F \end{bmatrix})} \quad (22)$$

where $K = \psi(X)^T \psi(X)$, $X = [X_s, X_t]$, all the $X_t$ are replaced by $\psi(X_t)$ and all the $X_s$ are replaced by $\psi(X_s)$ in $S_b, S, S_\omega, M_s, M, M_{st}$ and $M_{ts}$ in the kernelized version. Once the kernelized objective function Eq. (22) is obtained, we can simply solve it in the same way as the original objective function to compute $F_s$ and $F_t$. The Algorithm format of the proposed JSRDA can be seen on the website https://xdyangliu.github.io/JSRDA/ due to the limited space of the paper.

## IV. EXPERIMENTS

In this section, we evaluate our proposed approach on four public multi-view action datasets: the IXMAS action dataset [38], the Northwestern UCLA Multiview Action 3D (NUMA) dataset [39], the WVU action dataset [40] and the MuHAVi dataset [41].

We consider both cross-view and multi-view action recognition scenarios in this paper. The former one trains a classifier on one view (source view) and test it on the other view (target view), while the latter trains a classifier on $V-1$ views (source views) and test it on the remaining one view (target view). 1-Nearset Neighbor Classifier (NN) is adopted as the classifier. We adopt the improved dense trajectories (iDTs) [42] features with trajectory shape, HOG, HOF, MBHx, and MBHy as the low-level action video representation. The total length of the feature vector is 426. Then we adopt Locality-constrained Linear Coding (LLC) [43] scheme to represent the iDTs by 5 local bases, and the codebook size is set to be 2,000 for all training-testing partitions. Thus, the dimension of the encoded iDTs features is 2,000. To reduce the complexity when constructing the codebook, only 200 local iDTs are randomly selected from each video sequence according to the default setting in [28].

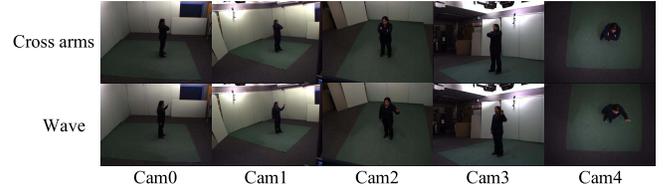

Fig. 3. Exemplar frames from the IXMAS dataset. Each row shows one action captured by five cameras.

For shared features learning, we fix noise probability $N_p = 0.6$ and number of layers $L = 1$ in all the experiments. For transferable dictionary learning, we set dictionary size $K = 1000$ and sparsity factor $\gamma = 50$. For distribution adaptation, we fix $\lambda = 1, \mu = 1$ for all the experiments, such that the distribution difference, subspace divergence and target view variance are treated as equally important, while the values of parameters $\beta$ and $k$ are chosen differently for various datasets.

For the action recognition task, it is hard to do eigendecomposition in the original data space since the dimensionality of data is high. Therefore, the experimental results are obtained using the RBF kernel in distribution adaptation, which is proved to be a good kernel for addressing the nonlinear problem by previous works [25], [44]. For fair comparison, we adopt the leave-one-action-class-out training strategy in [12] and [19]. At each time, only one action class is used for testing in the target view. In order to evaluate the effectiveness of our proposed approach, all the videos in this action class are excluded from the feature learning procedure including LLC encoding and the proposed JSRDA, while these action videos can be seen in training the classifiers. We report the classification accuracy by averaging all possible combinations for selecting testing action classes.

On Intel (R) CoreTM i7 system with 32GB RAM and un-optimized Matlab code excluding the process of iDTs extraction and LLC encoding, we get average run-time for training videos as 1.53 seconds and testing videos as 0.17 seconds for cross-view action recognition on the IXMAS dataset. For multi-view action recognition, we get average run-time for training videos as 4.99 seconds and testing videos as 0.99 seconds. Each video on the IXMAS dataset lasts for 4 seconds in average. We have released some extra experimental results and codes on the website https://xdyangliu.github.io/JSRDA/ due to the limited space of the paper.

### A. IXMAS Dataset

The IXMAS dataset has 1,650 action samples with 11 action classes recorded by 4 side view cameras and 1 top view camera. These actions are check watch, cross arms, get up, kick, pick up, punch, scratch head, sit down, turn around, walk and wave. Figure 3 shows some example frames.

*1) Cross-View Action Recognition:* In this experiment, we evaluate our proposed JSRDA approach for cross-view

TABLE I
CROSS-VIEW ACTION RECOGNITION RESULTS OF VARIOUS APPROACHES UNDER 20 COMBINATIONS OF SOURCE (TRAINING) AND TARGET (TESTING)
VIEWS ON THE IXMAS DATASET UNDER UNSUPERVISED MODE. 0 TO 4 DENOTE CAMERA0 TO CAMERA4 RESPECTIVELY

| Source\|Target | 0\|1 | 0\|2 | 0\|3 | 0\|4 | 1\|0 | 1\|2 | 1\|3 | 1\|4 | 2\|0 | 2\|1 | 2\|3 | 2\|4 | 3\|0 | 3\|1 | 3\|2 | 3\|4 | 4\|0 | 4\|1 | 4\|2 | 4\|3 | Ave. |
|---|---|---|---|---|---|---|---|---|---|---|---|---|---|---|---|---|---|---|---|---|---|
| Ulhaq et al. [22] | 65.1 | 64.2 | 66.5 | 65.9 | 64.6 | 64.5 | 69.4 | 69.4 | 67.2 | 65.9 | 67.3 | 67.5 | 71.4 | 63.5 | 70.0 | 62.1 | 68.3 | 66.6 | 67.7 | 66.2 | 66.7 |
| Rahmani et al. [23] | 92.7 | 80.3 | 83.9 | 55.2 | 95.5 | 80.6 | 86.4 | 47.0 | 82.7 | 83.6 | 83.6 | 75.5 | 85.8 | 85.2 | 84.9 | 44.2 | 56.0 | 53.0 | 79.0 | 52.4 | 74.1 |
| Liu et al. [19] | 79.9 | 76.8 | 76.8 | 74.8 | 81.2 | 75.8 | 78.0 | 70.4 | 79.6 | 76.6 | 79.8 | 72.8 | 73.0 | 74.1 | 74.4 | 66.9 | 82.0 | 68.3 | 74.0 | 71.1 | 75.3 |
| Zheng et al. [12] | 96.7 | 97.9 | 97.6 | 84.9 | 97.3 | 96.4 | 89.7 | 81.2 | 92.1 | 89.7 | 94.9 | 89.1 | 97.0 | 94.2 | 96.7 | 83.9 | 83.0 | 70.6 | 89.7 | 83.7 | 90.3 |
| Zheng et al. [14] | 99.1 | 90.9 | 88.7 | 95.5 | 97.8 | 91.2 | 78.4 | 88.4 | 99.4 | 97.6 | 91.2 | 100 | 87.6 | 98.2 | 99.4 | 95.4 | 87.3 | 87.8 | 92.1 | 90.0 | 92.8 |
| Zhu et al. [28] | 95.3 | 93.9 | 94.4 | 93.1 | 94.9 | 93.5 | 93.1 | 92.7 | 93.2 | 93.2 | 94.8 | 93.5 | 94.6 | 93.4 | 95.4 | 92.3 | 93.1 | 92.9 | 94.5 | 92.6 | 93.7 |
| Kong et al. [15] | 99.7 | 99.7 | 98.9 | 99.4 | **100** | 99.7 | 99.4 | 99.7 | **100** | 99.7 | **100** | 99.7 | **100** | **100** | **100** | **100** | 99.7 | **100** | **100** | **100** | 99.7 |
| Ours | **100** | **100** | **99.9** | **99.8** | **100** | **99.9** | **99.9** | **99.9** | **100** | **99.9** | **100** | **100** | 99.9 | **100** | 99.9 | 99.8 | **99.9** | 99.9 | **100** | 99.9 | **99.9** |

TABLE II
MULTI-VIEW ACTION RECOGNITION RESULTS ON THE IXMAS
DATASET. EACH COLUMN CORRESPONDS TO A TEST VIEW

| Methods | C0 | C1 | C2 | C3 | C4 | Ave. |
|---|---|---|---|---|---|---|
| Junejo et al. [6] | 74.8 | 74.5 | 74.8 | 70.6 | 61.2 | 71.2 |
| Rahmani et al. [23] | 78.4 | 78.0 | 80.7 | 75.8 | 57.8 | 74.1 |
| Liu and Shah [19] | 86.6 | 81.1 | 80.1 | 83.6 | 82.8 | 82.8 |
| Weinland et al. [46] | 86.7 | 89.9 | 86.4 | 87.6 | 66.4 | 83.4 |
| Yan et al. [45] | 91.2 | 87.7 | 82.1 | 81.5 | 79.1 | 84.3 |
| Ulhaq et al. [22] | 98.6 | 90.5 | 89.8 | 88.0 | 82.7 | 88.1 |
| Zhu et al. [28] | 94.6 | 96.8 | 97.4 | 95.2 | 95.9 | 95.9 |
| Zheng and Jiang [12] | 98.5 | 99.1 | 99.1 | **100** | 90.3 | 97.4 |
| Zheng et al. [14]-1 | 97.0 | **99.7** | 97.2 | 98.0 | 97.3 | 97.8 |
| Zheng et al. [14]-2 | 98.5 | 99.1 | 99.1 | **100** | 90.3 | 97.4 |
| Kong et al. [15] | **100** | **99.7** | **100** | **100** | 99.4 | **99.8** |
| No-JSRDA | 27.5 | 6.7 | 15.6 | 23.7 | 18.3 | 18.4 |
| No-TDL | 55.3 | 17.0 | 15.8 | 51.7 | 38.3 | 35.6 |
| No-DA | 91.8 | 92.6 | 93.6 | 92.9 | 91.4 | 92.5 |
| No-SFL | 97.1 | 96.3 | 98.1 | 97.0 | 95.7 | 96.8 |
| JSRDA | 99.8 | 99.3 | 99.8 | 99.8 | **99.5** | 99.6 |

action recognition on the IXMAS dataset. We compare our approaches with [12], [14], [15], [19], [22], [23], [28], and report recognition results in Table I. Our proposed approach achieves the best performance in 15 out of 20 combinations and the overall performance is significantly better than all the comparison approaches especially when the view difference is large (C4). In addition, our approach achieves nearly perfect performance on the IXMAS dataset with eight 100% accuracies, which verifies that our proposed approach is robust to viewpoint variations and can achieve good performance in cross-view action recognition with learned view-invariant representations.

*2) Multi-View Action Recognition:* In this experiment, we evaluate the performance of our proposed JSRDA approach for multi-view action recognition on the IXMAS dataset and make comparisons with existing approaches [6], [12], [14], [15], [19], [22], [23], [28], [45], [46]. The importance of shared feature learning, transferable dictionary learning and distribution adaptation are also evaluated. The No-JSRDA represents the results of 1-NN classifier without using our approach, while the No-SFL, No-TDL and No-DA represent the results of JSRDA method without shared features learning, transferable dictionary learning and distribution adaptation, respectively.

Table II shows that our proposed approach JSRDA achieves competitive recognition performance compared with other approaches. Although Kong *et al.* [15] achieves nearly perfect performance, our approach achieves comparable performance and achieves slightly better accuracies in C4 where C4 is the top view camera and there exists larger domain divergence between the top view and other side views. In addition, the overall performance of our method (99.6%) is comparable to Kong *et al.* [15] (99.8%). To be noticed, Zheng *et al.* [14] achieves satisfactory performance in C0, C1, C2 and C3 but the performance drops a lot when the target view is C4. On the contrary, our approach can still achieve satisfactory performance even when the target view is C4. These validate that our learned action representations is view-invariant and generalizes well across views even when the view difference is large.

The No-JSRDA performs very poorly and the recognition accuracy for most tasks is less than 20% due to the existence of domain divergence across views. Our proposed approach outperforms No-SFL, which verifies the effectiveness of the share features. Without shared features, No-SFL only utilize the private features which are not discriminative enough as some motion information only exist in one view and cannot be shared across views. There is a large margin between the accuracies of Ours and that of the No-TDL, which demonstrates that transferable dictionary learning can encourage the samples from different views to have the same sparse representation and thus reduce the domain divergence effectively. In addition, the accuracy gap between Ours and No-DA suggests the benefit of distribution adaptation for learning more robust view-invariant representations that generalize well across views especially when the view difference is large. More importantly, the shared features learning (SFL), transferable dictionary learning (TDL) and distribution adaptation (DA) are complementary and indeed encourage us to learn view-invariant features hierarchically.

*3) Parameter Analysis:* We analyse the sensitivity of parameters $N_p$, $L$, $K$, $\beta$, $k$ and $T$ in this experiment while fixing $\lambda = 1$, $\mu = 1$ and $\gamma = 50$. We conduct experiments on the multi-view action recognition task C0. The results of other multi-view action recognition tasks or cross-view action recognition tasks are not given here as it shows similar results to the C0.

$N_p$ is the corruption probability in shared features learning stage, we evaluate its performance given values $N_p \in \{0, 0.1, 0.2, 0.3, 0.4, 0.5, 0.6, 0.7, 0.8, 0.9\}$. Results in Figure 4(a) indicate that the performance increases if we increase the noise probability $N_p$. When $N_p$ is lower than 0.3, the performance is poor. As the random corruption is essentially adding dropout regularization, too lower corruption probability cannot guarantee obtaining the informative corrupted data which is different from raw data. When $N_p$ exceeds 0.3, the accuracy increases a lot and reach its optimum value when $N_p = 0.6$. If we go on increasing the values of $N_p$, the performance becomes decreasing. The underlying reason is that adding too much noise in raw data ($N_p > 0.6$) reduces the amount of shared information between views. Thus, the discriminative power of shared features is decreased and lead to a relatively lower recognition accuracy. Considering these issues, we use $N_p = 0.6$ in this work.

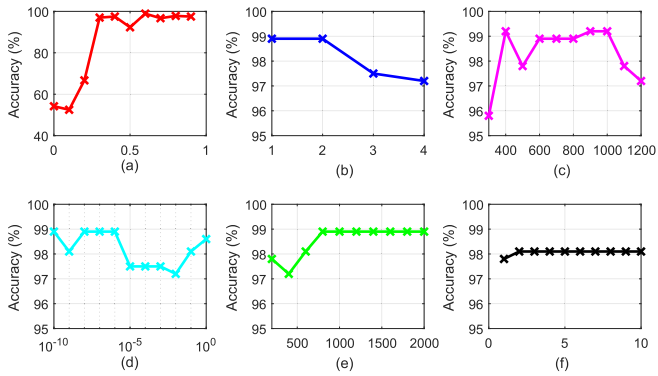

Fig. 4. Performance analysis of the JSRDA on IXMAS dataset with various values of parameters $N_p$, $L$, $K$, $\beta$, $k$ and $T$. (a) Value of parameter $N_p$. (b) Number of layers $L$. (c) Value of parameter $K$. (d) Value of parameter $\beta$. (e) Value of parameter $k$. (f) Number of iterations $T$.

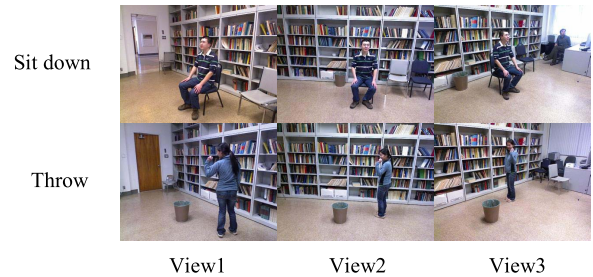

Fig. 5. Exemplar frames from the Northwestern UCLA dataset. Action classes: Sit down and Throw.

$L$ is the number of layers in mSDA, we evaluate its performance given values $L \in \{1, 2, 3, 4\}$. From Figure 4(b), we can see that the performance of $L = 1$ and $L = 2$ is the same, then the performance decreases when $L > 2$ due to the redundant information results from multiple layers. Considering the extra training time using multiple layers, we use $L = 1$ in this work.

$K$ is the dictionary size in transferable dictionary learning stage, we fix the sparsity factor $\gamma = 50$ and vary the dictionary size from 300 to 1200 in multiples of 100 as in the range $\{300, 400, 500, 600, 700, 800, 900, 1000, 1100, 1200\}$ due to the fact that the dictionary size must be larger than the number of samples in the IXMAS dataset of each view to guarantee the sparsity. From Figure 4 (c), we observe that the performance increases as the dictionary size increases and keep stable from $K = 600$ to $K = 1000$. However, when the dictionary size $K$ is too large ($K > 1000$), the redundancy in dictionaries will affect the sparse representation of action samples and thus performance may decrease. Therefore, we choose $K = 1000$ in this work considering both the sparsity and the performance.

$\beta$ is the trade-off parameters of intra-class and inter-class variance of source view. A large range of $\beta$ ($\beta \in [10^{-10}, 1]$) are selected to evaluate its effect on the overall performance. If $\beta$ is too small, the class information of source view is not considered. If $\beta$ is too large, the classifier may be overfit to the source view. As can be seen from Figure 4 (d), the performance is stable and good when $\beta$ is neither too small nor too big. To make a balance between the class information and the overfit problem, we use $\beta = 10^{-6}$ in this work.

$k$ is the dimension of the learned view-invariant features, which is essentially the number of the chosen eigenvectors in eigendecomposition at distribution adaptation stage. We illustrate the relationship between various $k$ and the overall accuracy given values $k \in \{200, 400, 600, 800, 1000, 1200, 1400, 1600, 1800, 2000\}$. From Figure 4 (e), we can observe that the performance becomes stable when $k$ is larger than a certain value (800). This is because much information of feature representation may lose in eigendecomposition process when $k$ is too small. Thus, discriminability of the feature is not enough and the overall performance is unsatisfactory. But when $k$ exceeds a certain value, information can be reserved well and $k$ has little impact on the overall performance except for the computational complexity. Although we can choose $k \in \{800, 2000\}$ to obtain view-invariant representations due to their good performance, we use $k = 1000$ in this work considering the time efficiency.

The convergence of the proposed method is also verified by analyzing the number of iterations $T$. As can be seen from Figure 4 (f), the accuracy converges to the optimum value only after 2 iterations and then keep stable. Therefore, we use $T = 5$ in this work.

### B. Northwestern UCLA Multiview Action 3D Dataset

The NUMA dataset has 1,509 action samples with 10 action classes captured by 3 Kinect cameras in 5 environments. These actions includes pick up with one hand, pick up with two hands, drop trash, walk around, sit down, stand up, donning, doffing, throw and carry. Figure 5 shows exemplar frames of four action classes taken by three cameras.

*1) Cross-View Action Recognition:* In this experiment, we evaluate our proposed JSRDA approach for cross-view action recognition on the NUMA dataset. Our method is compared with [24], [28], and [47]–[50]. Results in Table III show that our proposed JSRDA approach achieves the best performance in all combinations and outperforms all the comparison approaches by a significantly large accuracy margin. In addition, our approach achieves nearly perfect performance on the NUMA dataset with two 100% accuracies and four nearly perfect accuracies (99.9%), which verifies that our proposed approach is robust to viewpoint variations and can achieve good performance in cross-view action recognition with learned view-invariant representations. Compared with [48] and [49], our better performance may be due to the utilization of unlabeled target view data. These data contributes to cope with the data-distribution mismatch. The explanation for the better performance of ours than [24] and [50] may be the lack of strong manifold structure on these cross-view datasets. Approach [47] assumes that there exists a common subspace where the modality gap between datasets can be reduced effectively, but this assumption is invalid when the view difference is large. Such remarkable improvements demonstrate the benefit of using both shared and private features for modeling cross-view feature representation, and learning a set of projections that project the source and target domains into respective subspaces without the manifold structure assumption.

TABLE III
CROSS-VIEW ACTION RECOGNITION RESULTS OF VARIOUS APPROACHES ON THE NUMA DATASET UNDER UNSUPERVISED MODE. EACH ROW CORRESPONDS TO A TRAINING VIEW AND EACH COLUMN A TESTING VIEW. 0 TO 2 DENOTE VIEW1 TO VIEW3 RESPECTIVELY

| Source\|Target | 0\|1 | 0\|2 | 1\|0 | 1\|2 | 2\|0 | 2\|1 | Ave. |
|---|---|---|---|---|---|---|---|
| Wang et al. [24] | 40.8 | 49.8 | 48.4 | 51.3 | 47.8 | 40.7 | 46.7 |
| Kulis et al. [48] | 41.0 | 49.6 | 47.9 | 50.2 | 47.0 | 40.6 | 46.1 |
| Hoffman et al. [49] | 39.6 | 51.8 | 46.0 | 52.1 | 46.4 | 38.8 | 45.8 |
| Li et al. [50] | 42.8 | 53.7 | 47.5 | 53.5 | 47.9 | 42.7 | 48.0 |
| Sui et al. [47] | 43.1 | 56.5 | 49.6 | 57.1 | 49.0 | 43.7 | 49.8 |
| Zhu et al. [28] | 90.0 | 90.0 | 90.1 | 90.0 | 90.0 | 90.1 | 90.0 |
| Our | 99.9 | 99.9 | 100 | 99.9 | 99.9 | 100 | 99.9 |

TABLE IV
MULTI-VIEW ACTION RECOGNITION RESULTS ON THE NUMA DATASET. EACH COLUMN CORRESPONDS TO A TEST VIEW. THE SYMBOL 'N/A' DENOTES THAT THE RESULT IS NOT REPORTED IN THE PUBLISHED PAPER

| Methods | C0 | C1 | C2 | Ave. |
|---|---|---|---|---|
| Sui et al. [47] | 49.3 | 43.4 | 56.8 | 49.8 |
| Kong et al. [15] | N/A | N/A | N/A | 72.5 |
| Wang et al. [39] | N/A | N/A | N/A | 73.3 |
| Rahmani et al. [23] | N/A | N/A | N/A | 78.1 |
| Zhu et al. [28] | 90.2 | 90.2 | 90.3 | 90.2 |
| Rahmani et al. [51] | N/A | N/A | N/A | 92.0 |
| Liu et al. [52] | N/A | N/A | N/A | 92.6 |
| No-JSRDA | 12.5 | 8.8 | 11.8 | 11.0 |
| No-TDL | 14.3 | 11.0 | 11.0 | 12.1 |
| No-DA | 84.8 | 84.1 | 84.7 | 84.5 |
| No-SFL | 89.5 | 89.5 | 88.6 | 89.2 |
| JSRDA | 100 | 100 | 99.9 | 99.9 |

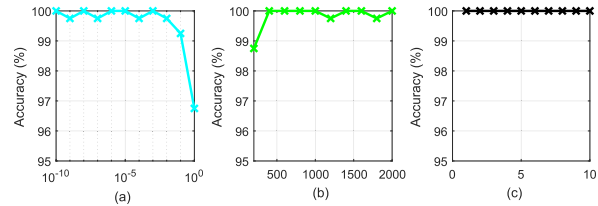

Fig. 6. Performance analysis of the JSRDA on NUMA dataset with various values of parameters $\beta$, $k$ and $T$. (a) Value of parameter $\beta$. (b) Value of parameter k. (c) Number of iterations T.

*2) Multi-View Action Recognition:* In this experiment, we evaluate the performance of our proposed JSRDA for multi-view action recognition on the NUMA dataset. Our approach is compared with [15], [23], [28], [39], [47], [51], and [52]. As most of the published papers only report the average accuracies in the multi-view scenario, we only quote the given accuracies in Table IV. The importance of shared features learning (SFL), transferable dictionary learning (TDL) and distribution adaptation (DA) are also evaluated.

Table IV shows that our proposed JSRDA achieves nearly perfect performance in the NUMA dataset with two 100% accuracies. Compared with other approaches, ours performs significantly better with a large accuracy margin. These demonstrate that our proposed JSRDA can learn robust and discriminative view-invariant representations for multi-view action recognition. Without the JSRDA, the No-JSRDA performs very poorly due to the existence of domain divergence across views. Thus, directly applying the classifier trained on one view to another view will degrade the performance. The No-SFL performs worse than the JSRDA demonstrates the effectiveness of shared features which are complementary to the private features. A large accuracy margin exists between ours and the No-TDL, which shows that transferable dictionary learning is a key stage for reduce the domain divergence across views by encouraging the samples from different views to have the same sparse representations. In addition, the accuracy gap between Ours and No-DA suggests the benefit of distribution adaptation for addressing the performance degradation problem caused by the large view difference. More importantly, the shared features learning (SFL), transferable dictionary learning (TDL) and distribution adaptation (DA) are complementary and indeed encourage us to learn view-invariant representations hierarchically.

*3) Parameter Analysis:* The sensitivity of parameters $\beta$, $k$ and $T$ are evaluated while fixing other parameters $\lambda = 1$, $\mu = 1$, $\gamma = 50$, $N_p = 0.6$, $L = 1$ and $K = 1000$ according to the sensitivity analysis results from the IXMAS dataset. We conduct experiments on the multi-view action recognition task C0.

$\beta$ is the trade-off parameters of intra-class and inter-class variance of source view. A large range of $\beta$ ($\beta \in [10^{-10}, 1]$) are selected to evaluate its effect on the overall performance. As can be seen from Figure 6 (a), our approach is insensitive to the parameter $\beta$ with small accuracy variation (0.25%) when $\beta$ is small. When $\beta$ is too large, the classifier may be overfit to the source view and thus the performance degrades. To make a balance between the class information and the overfit problem, we use $\beta = 10^{-6}$ in this work.

$k$ is the dimension of the learned view-invariant features. We illustrate the relationship between various $k$ and the overall accuracy given values $k \in \{200, 400, 600, 800, 1000, 1200, 1400, 1600, 1800, 2000\}$. From Figure 6 (b), we can observe that we can choose $k \in \{400, 1600\}$ to obtain view-invariant features due to their good performance. We make a compromise between the time efficiency and performance and use $k = 1000$ in this work.

The convergence of the proposed method is also verified by analysing the number of iterations $T$. As can be seen from Figure 6 (c), the accuracy converges to the optimum value within only 1 iteration and then keep stable. Therefore, we use $T = 5$ in this work.

*C. WVU Action Dataset*

The WVU dataset is collected from a network of 8 embedded cameras with 12 action classes and each action

TABLE V

CROSS-VIEW ACTION RECOGNITION RESULTS OF VARIOUS APPROACHES ON THE WVU DATASET UNDER UNSUPERVISED MODE. EACH ROW CORRESPONDS TO A TRAINING VIEW AND EACH COLUMN A TESTING VIEW. C0 TO C7 DENOTE VIEW1 TO VIEW8 RESPECTIVELY. THE THREE ACCURACY NUMBERS IN THE BRACKET ARE THE AVERAGE RECOGNITION ACCURACIES OF [14], [20], AND OUR PROPOSED APPROACH RESPECTIVELY

|    | C0 | C1 | C2 | C3 | C4 | C5 | C6 | C7 |
|----|----|----|----|----|----|----|----|----|
| C0 | NA | (92.5, **100**, 98.9) | (89.3, **100**, 98.6) | (90.2, **100**, 97.0) | (90.9, **100**, 98.3) | (88.2, **99.6**, 98.7) | (90.7, **99.8**, 92.1) | (91.6, **100**, 99.1) |
| C1 | (87.3, **100**, 98.4) | NA | (86.8, **100**, 98.6) | (89.3, **100**, 97.7) | (84.3, **99.8**, 97.5) | (92.5, **98.6**, 98.4) | (86.6, **99.6**, 92.1) | (89.5, **100**, 98.6) |
| C2 | (88.9, 91.9, **98.9**) | (90.7, 81.8, **99.2**) | NA | (89.8, 89.9, **97.5**) | (85.0, 90.5, **97.6**) | (90.5, 91.1, **98.8**) | (89.8, 90.1, **90.5**) | (92.3, 89.7, **98.9**) |
| C3 | (86.1, **100**, 98.4) | (92.3, **99.6**, 98.3) | (85.7, **99.6**, 98.2) | NA | (86.1, **98.8**, 94.1) | (90.5, 88.4, **98.4**) | (89.8, **97.1**, 91.1) | (92.3, **99.8**, 98.6) |
| C4 | (91.1, 98.2, **99.4**) | (87.7, 90.1, **99.1**) | (86.4, 94.0, **99.0**) | (92.7, **98.6**, 98.3) | NA | (91.4, **99.2**, 98.2) | (86.6, 90.0, **90.1**) | (91.8, 94.8, **99.3**) |
| C5 | (90.0, 93.0, **99.0**) | (92.0, 89.0, **99.1**) | (90.0, 81.4, **99.0**) | (90.0, 89.7, **98.0**) | (90.5, 90.1, **95.8**) | NA | (89.3, 82.6, **91.3**) | (90.2, 76.4, **99.1**) |
| C6 | (88.0, 81.4, **93.3**) | (89.8, 74.2, **94.5**) | (89.8, 83.8, **91.3**) | (90.0, 81.2, **93.9**) | (83.6, 81.6, **90.4**) | (89.8, 81.6, **91.0**) | NA | (91.6, 82.2, **92.2**) |
| C7 | (90.0, **98.8**, 98.2) | (91.6, 91.3, **98.5**) | (88.4, 97.3, **98.6**) | (92.0, 90.9, **98.3**) | (86.4, 91.9, **97.3**) | (90.2, 89.4, **98.1**) | (88.6, **96.5**, 90.3) | NA |
| Ave. | (88.8, 94.7, **97.9**) | (90.9, 89.6, **98.2**) | (88.1, 93.3, **97.6**) | (90.6, 93.0, **97.2**) | (86.7, **98.9**, 95.9) | (90.4, 92.6, **97.4**) | (88.3, **93.6**, 90.9) | (91.2, 91.8, **98.0**) |

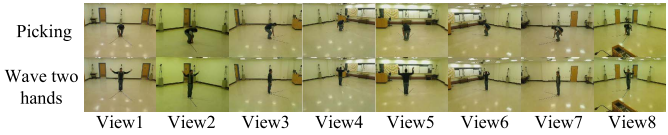

Fig. 7. Exemplar frames from the WVU dataset. Each row shows one action viewed across eight camera views.

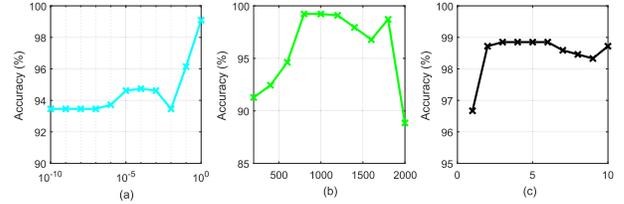

Fig. 8. Performance analysis of the JSRDA on WVU dataset with various values of parameters $\beta$, $k$ and $T$. (a) Value of parameter $\beta$. (b) Value of parameter k. (c) Number of iterations T.

has 65 video samples. There are a total of 6,240 video samples in this dataset, which is a relatively large multi-view action dataset compared with the IXMAS and the NUMA datasets. These actions includes standing still, nodding head, clapping, waving one hand, waving two hands, punching, jogging, jumping jack, kicking, picking, throwing and bowling. Figure 7 shows exemplar frames of two action classes taken by eight cameras.

*1) Cross-View Action Recognition:* In this experiment, we evaluate our proposed JSRDA approach for cross-view action recognition on the WVU dataset. Our method is compared with two approaches in [14] and [20]. Results in Table V show that our approach achieves similar performance compared with [14]. Some accuracies in [14] are better demonstrates the effectiveness of transferable dictionary learning proposed in [14], but when the distribution divergence across views are large, transferable dictionary learning cannot address the cross-view problem well without considering distribution adaptation. For example, when the source view is C2, C5 and C6, our proposed approach outperforms all the pairwise views by a large margin compared with the approach [14]. In addition, we achieve the best average accuracies in 6 out of 8 tasks including C0, C1, C2, C3, C5 and C7. These verify that our approach can effectively address the cross-view action recognition problem by learning view-invariant representations using the novel JSRDA framework.

*2) Parameter Analysis:* We also evaluate the sensitivity of our approach to parameters $\beta$, $k$ and $T$ while fixing other parameters $\lambda = 1$, $\mu = 1$, $\gamma = 50$, $N_p = 0.6$, $L = 1$ and $K = 1000$ according to the sensitivity analysis results from the IXMAS dataset and the NUMA dataset. We conduct experiments on the cross-view action recognition task C0→C1.

$\beta$ is the trade-off parameters of intra-class and inter-class variance of source view. A large range of $\beta$ ($\beta \in [10^{-10}, 1]$) are selected to evaluate its effect on the overall performance. As can be seen from Figure 8 (a), our approach is insensitive to the parameter $\beta$ when it is small, and achieves the best performance when $\beta = 1$. Therefore, we use $\beta = 1$ in this work.

$k$ is the dimension of the learned view-invariant features. We illustrate the relationship between various $k$ and the overall accuracy given values $k \in \{200, 400, 600, 800, 1000, 1200, 1400, 1600, 1800, 2000\}$. From Figure 8 (b), we can observe that we can choose $k \in \{800, 1200\}$ to obtain view-invariant features due to their good performance. We make a compromise between the time efficiency and performance and use $k = 1000$ in this work.

The convergence of the proposed method is also verified by analysing the number of iterations $T$. As can be seen from Figure 8 (c), the accuracy converges to the optimum value within 5 iterations. Therefore, we use $T = 5$ in this work.

### D. MuHAVi Dataset

The MuHAVi dataset [41] contains 17 human action classes: WalkTurnBack, RunStop, Punch, Kick, ShotGunCollapse, PullHeavyObject, PickupThrowObject, WalkFall, LookInCar, CrawlOnKnees, WaveArms, DrawGraffiti, JumpOverFence, DrunkWalk, ClimbLadder, SmashObject and JumpOverGap. Each action video is performed by 7 actors and recorded using 8 CCTV Schwan cameras located at 4 sides and 4 corners of a rectangular platform. To reduce the computational burden and have a fair comparison with other works, we follow [14] and [53] to choose the action videos captured by four cameras (i.e. two side cameras and two corner cameras) in our experiments. Figure 9 shows exemplar frames of two action classes taken by four cameras.

Table VI shows the recognition accuracies of our proposed JSRDA for cross-view action recognition. Although both WSCDD [28] and un-RLTDL [14] are based on transferable

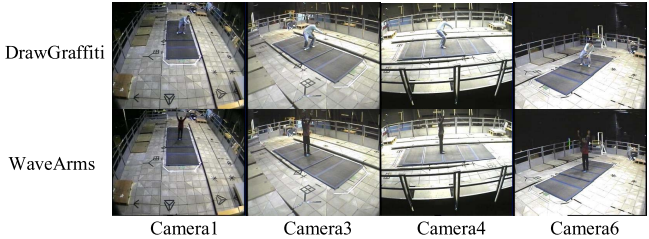

Fig. 9. Exemplar frames from the MuHAVi dataset. Action classes: DrawGraffiti and WaveArms.

TABLE VI

CROSS-VIEW ACTION RECOGNITION RESULTS OF VARIOUS APPROACHES ON THE MUHAVI DATASET UNDER UNSUPERVISED MODE. EACH ROW CORRESPONDS TO A TRAINING VIEW AND EACH COLUMN A TESTING VIEW. THE THREE ACCURACY NUMBERS IN THE BRACKET ARE THE AVERAGE RECOGNITION ACCURACIES OF WSCDD [28], UN-RLTDL [14] AND OUR PROPOSED APPROACH RESPECTIVELY

|  | C1 | C3 | C4 | C6 |
|---|---|---|---|---|
| C1 | NA | (98.1, 99.8, **99.9**) | (97.6, 96.6, **99.9**) | (98.9, 99.8, **100**) |
| C3 | (97.8, 98.3, **99.9**) | NA | (98.9, 98.3, **99.9**) | (98.0, 99.8, **100**) |
| C4 | (97.5, 98.3, **99.9**) | (99.2, 97.5, **99.8**) | NA | (97.4, 99.2, **100**) |
| C6 | (99.0, 95.0, **99.9**) | (98.2, 89.1, **99.9**) | (97.5, 92.4, **99.9**) | NA |
| Ave. | (98.1, 98.6, **99.9**) | (98.5, 98.9, **99.8**) | (98.0, 97.8, **99.9**) | (98.1, 99.8, **100**) |

TABLE VII

MULTI-VIEW ACTION RECOGNITION RESULTS OF VARIOUS APPROACHES ON THE MUHAVI DATASET UNDER UNSUPERVISED MODE. EACH COLUMN CORRESPONDS TO A TEST VIEW

| Methods | C1 | C3 | C4 | C6 | Ave. |
|---|---|---|---|---|---|
| Chang et al. [54] | 93.3 | 92.4 | 93.3 | 95.8 | 93.7 |
| Yu et al. [55] | 91.6 | 94.1 | 95.8 | 95.8 | 94.3 |
| Wu et al. [53] | 96.6 | 93.3 | 94.1 | 94.1 | 94.5 |
| Zheng et al. [14] | 96.6 | 97.5 | **99.8** | 99.8 | 98.5 |
| Zhu et al. [28] | 99.2 | 100 | 98.8 | 100 | 99.5 |
| JSRDA_actor | **99.8** | 99.8 | 99.5 | 99.9 | 99.7 |
| JSRDA | 99.7 | **100** | 99.8 | **100** | **99.8** |

TABLE VIII

PERFORMANCE COMPARISON OF MULTI-VIEW ACTION RECOGNITION TASK C4 ON THE IXMAS DATASET FOR DIFFERENT COMBINATIONS OF FEATURES AND CODEBOOK SIZES. D DENOTES THE CODEBOOK SIZE

| Feature type | D=1000 | D=2000 | D=4000 | D=8000 | Ave. |
|---|---|---|---|---|---|
| Tra+HOG+HOF | 96.2 | 98.4 | 97.8 | 98.7 | 97.7 |
| Tra+MBHx+MBHy | 96.6 | 97.0 | 98.2 | 97.4 | 97.3 |
| HOF+MBHx+MBHy | 96.5 | 98.7 | 98.6 | 98.8 | 98.1 |
| HOG+MBHx+MBHy | 97.0 | 98.9 | 97.7 | 99.4 | 98.2 |
| Tra+HOG+MBHx+MBHy | 97.9 | 98.3 | 97.8 | 99.0 | 98.2 |
| Tra+HOF+MBHx+MBHy | 96.8 | 98.1 | 98.2 | 98.1 | 97.8 |
| HOG+HOF+MBHx+MBHy | 97.6 | 97.6 | 98.6 | 98.7 | 98.1 |
| Tra+HOG+HOF+MBHx+MBHy | 97.1 | 99.0 | 98.3 | 99.0 | **98.4** |

dictionary learning, JSRDA achieves better performance than WSCDD and un-RLTDL due to the benefits of shared features and distribution adaptation. This demonstrates that JSRDA is robust to viewpoint variations and can achieve satisfactory performance in cross-view action recognition.

We also evaluate our approach for multi-view action recognition on the MuHAVi dataset and the results can be seen in Table VII. Although Zheng et al. [14] achieves good performance in C4 and C6, its performance degrades when the target view is C1 and C3 due to the existence of large view difference. However, JSRDA can still achieve good performance even when the view difference is large and the overall performance is better than other approaches [14], [28], [53]–[55]. This illustrates that JSRDA can learn robust and discriminative view-invariant representations for multi-view action recognition even with large view difference.

To evaluate whether our method can generalize both the view and the action class, we use the leave-one-actor-out strategy for training and testing which means that each time one actor is excluded from both training and testing procedure. We report the classification accuracy by averaging all possible combinations for excluding actors. We conduct experiments for multi-view action recognition task on the MuHAVi dataset.

From Table VII, we can see that the performance of our method using leave-one-actor-out strategy (JSRDA_actor) is comparable to that of our method using leave-one-action-class-out strategy (JSRDA). This verifies that our method can generalize both the view and the action class.

### E. Impact of Features Extraction Parameters

To have a more detailed study on how well our method behaves for different choices of features, we use some possible combinations of features (trajectory shape, HOG, HOF, MBHx and MBHy) to form the iDTs to evaluate our method. To have a more detailed study on how well our method behaves for different choices of extraction parameters, we conduct experiments for multi-view action recognition task C4 on the IXMAS dataset with different codebook sizes while keeping other parameters unchanged. The results can be seen in Table VIII. We can see that the average accuracy is the best when we combine trajectory shape (Tra), HOG, HOF, MBHx and MBHy as the video feature. Although the accuracy of $D = 2000$ is the same as that of $D = 8000$, we make a compromise between performance and computational complexity and choose 2000 as the codebook size in our experiments.

## V. CONCLUSION

In this paper, we propose a novel view-invariant representation learning approach for cross-view action recognition. Our approach incorporates shared features learning, transferable dictionary learning and distribution adaptation into a unified framework and learns view-invariant representations hierarchically. A sample affinity matrix is incorporated into the marginalized stacked denoising Autoencoder (mSDA) to learn shared features, and the shared features are combined with the private features to obtain new informative feature representation. Then, a transferable dictionary pair is learned simultaneously from pairs of videos taken at different views to encourage each action video across views to have the same sparse representation. To address the problem of large view difference, a novel unsupervised distribution adaptation method is proposed to reduce the difference in both the marginal distribution and conditional distribution across views by learning a set of projections that project the source and target views data into respective low-dimensional subspaces. Finally, the view-invariant representations of action videos from different views are obtained in their respective subspaces.

Extensive experiments on the IXMAS, NUMA, WVU and MuHAVi datasets show that our approach outperforms state-of-the-art approaches for both cross-view and multi-view action recognition.

## APPENDIX A
### THE FORMULATION OF EACH TERM IN (7)

$F_s$ and $F$ are defined in Eq. (23),

$$F_s = \begin{bmatrix} F_{s,1} \\ \cdots \\ F_{s,p} \end{bmatrix}, \quad X_s = [X_{s,1}, \cdots, X_{s,p}]$$

$$F = \begin{bmatrix} F_t \\ \cdots \\ F_t \end{bmatrix} \text{ is obtained by replicating } F_t \text{ p times.}$$

$X = [X_t, \cdots, X_t]$ is obtained by replicating $X_t$ p times. (23)

$M_s$ is defined in Eq. (24),

$$M_s = X_s(L_s + \sum_{c=1}^{C} L_s^{(c)})X_s^{\mathrm{T}}, \quad L_s = [L_{s,1}, \cdots, L_{s,p}],$$

$$L_{s,i} = \frac{1}{N_{s,i}^2} 1_{s,i} 1_{s,i}^{\mathrm{T}}, \quad 1_{s,i} \in \mathbb{R}^{N_{s,i} \times 1},$$

$$L_s^{(c)} = [L_{s,1}^{(c)}, \cdots, L_{s,p}^{(c)}], \quad i \in \{1, \cdots, p\},$$

$$(L_{s,i}^{(c)})_{m,n} = \begin{cases} \frac{1}{(N_{s,i}^{(c)})^2}, & \mathbf{x}_m, \mathbf{x}_n \in X_{s,i}^{(c)} \\ 0, & \text{otherwise} \end{cases} \quad (24)$$

$M_{st}$ is defined in Eq. (25),

$$M_{st} = X_s(L_{st} + \sum_{c=1}^{C} L_{st}^{(c)})X^{\mathrm{T}},$$

$$L_{st} = [L_{st,1}, \cdots, L_{st,p}],$$

$$L_{st,i} = \frac{1}{N_{s,i} N_t} 1_{s,i} 1_t^{\mathrm{T}}, \quad 1_{s,i} \in \mathbb{R}^{N_{s,i} \times 1}, \quad 1_t \in \mathbb{R}^{N_t \times 1}$$

$$L_{st}^{(c)} = [L_{st,1}^{(c)}, \cdots, L_{st,p}^{(c)}], \quad i \in \{1, \cdots, p\},$$

$$(L_{st,i}^{(c)})_{m,n} = \begin{cases} -\frac{1}{N_{s,i}^{(c)} N_t^{(c)}}, & \mathbf{x}_m \in X_{s,i}^{(c)}, \mathbf{x}_n \in X_t^{(c)} \\ 0, & \text{otherwise} \end{cases} \quad (25)$$

$M_{ts}$ is defined in Eq. (26),

$$M_{ts} = X(L_{ts} + \sum_{c=1}^{C} L_{ts}^{(c)})X_s^{\mathrm{T}}, \quad L_{ts} = [L_{ts,1}, \cdots, L_{ts,p}],$$

$$L_{ts,i} = \frac{1}{N_t N_{s,i}} 1_t 1_{s,i}^{\mathrm{T}}, \quad 1_t \in \mathbb{R}^{N_t \times 1}, \quad 1_{s,i} \in \mathbb{R}^{N_{s,i} \times 1}$$

$$L_{ts}^{(c)} = [L_{ts,1}^{(c)}, \cdots, L_{ts,p}^{(c)}], \quad i \in \{1, \cdots, p\},$$

$$(L_{ts,i}^{(c)})_{m,n} = \begin{cases} -\frac{1}{N_t^{(c)} N_{s,i}^{(c)}}, & \mathbf{x}_m \in X_t^{(c)}, \mathbf{x}_n \in X_{s,i}^{(c)} \\ 0, & \text{otherwise} \end{cases} \quad (26)$$

$M$ is defined in Eq. (27),

$$M = X(L + \sum_{c=1}^{C} L^{(c)})X^{\mathrm{T}},$$

$L = [L_t, \cdots, L_t]$ is obtained by replicating $L_t$ p times.

$$L_t = \frac{1}{N_t^2} 1_t 1_t^{\mathrm{T}}, \quad 1_t \in \mathbb{R}^{N_t \times 1},$$

$L^{(c)} = [L_t^{(c)}, \cdots, L_t^{(c)}]$ is obtained by replicating $L_t^{(c)}$ p times.

$$(L_t^{(c)})_{m,n} = \begin{cases} \frac{1}{(N_t^{(c)})^2}, & \mathbf{x}_m, \mathbf{x}_n \in X_t^{(c)} \\ 0, & \text{otherwise} \end{cases} \quad (27)$$

where $1_{s,i}$ and $1_t$ are the column vectors with all ones.

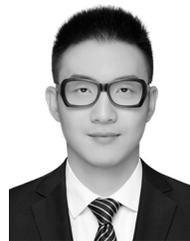

**Yang Liu** received the B.S. degree from the School of Information Engineering, Chang'an University, Xi'an, China, in 2014. He is currently pursuing the Ph.D. degree with the School of Telecommunications Engineering, Xidian University, Xi'an. His research interests include cross-domain action recognition and transfer learning.

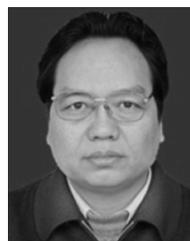

**Zhaoyang Lu** (SM'14) received the bachelor's, master's, and Ph.D. degrees in communication and information systems from Xidian University, Xi'an, China, in 1982, 1985, and 1990, respectively. He is currently a Full Professor with the School of Telecommunications Engineering, Xidian University. His current research interests include image matching and recognition and video content analysis and understanding. He has authored or co-authored over 100 papers. He holds over 10 patents in the field of pattern recognition and image processing.

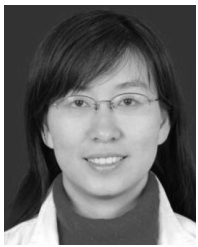

**Jing Li** (M'14) received the Ph.D. degree in control theory and engineering from Northwestern Polytechnical University, Xi'an, China, in 2008. From 2004 to 2005, she was a Visiting Scholar with the National Laboratory of Pattern Recognition, Beijing, China. In 2008, she was a Research Assistant with the Department of Computing, The Hong Kong Polytechnic University. She was a Visiting Scholar with the University of Delaware, USA, from 2013 to 2014. She is currently an Associate Professor with the School of Telecommunications Engineering, Xidian University, Xi'an, where she is also the Leader of the Intelligent Signal Processing and Pattern Recognition Laboratory. She has published over 50 research papers in international journals and conference proceedings in the areas of computer vision and pattern recognition. Her research interests include image registration, matching and retrieval, and video content analysis and understanding.

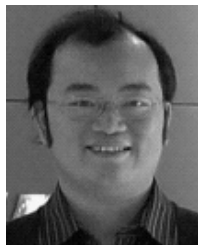

**Tao Yang** (M'13) received the Ph.D. degree in control theory and engineering from Northwestern Polytechnical University, Xi'an, China, in 2008. He was a Visiting Scholar with the Intelligent Video Surveillance Group, National Laboratory of Pattern Recognition, Beijing, China, from 2004 to 2005, and Microsoft Research Asia. He was a Research Intern with the FX Palo Alto Laboratory (FXPAL), Palo Alto, CA, USA, from 2006 to 2007. He was a Post-Doctoral Fellow with the Shaanxi Provincial Key Laboratory of Speech and Image Information Processing, Northwestern Polytechnical University, from 2008 to 2010. He was a Visiting Scholar with the University of Delaware, USA, from 2013 to 2014. He is currently a Full Professor with the School of Computer Science, Northwestern Polytechnical University. His research interests include video content analysis and understanding and image and video registration. He has published over 50 research papers in these fields. He is serving as a reviewer to numerous international journals, conferences, and funding agencies.